\documentclass[10pt,twocolumn,letterpaper]{article}

\usepackage{cvpr}
\usepackage{times}
\usepackage{epsfig}
\usepackage{graphicx}
\usepackage{amsmath}
\usepackage{amssymb}

\usepackage[pagebackref=true,breaklinks=true,letterpaper=true,colorlinks,bookmarks=false]{hyperref}

\cvprfinalcopy % *** Uncomment this line for the final submission

\def\cvprPaperID{8084} % *** Enter the CVPR Paper ID here

\ifcvprfinal\fi
\begin{document}

%%%%%%%%% TITLE
\title{AI Online Filters to Real World Image Recognition}

\author{Hai Xiao\\
Stanford University\\
{\tt\small haixiao@stanford.edu}
\and
Jin Shang \\
Sunnyvale, USA \\
{\tt\small shangjin@gmail.com}
\and
Mengyuan Huang\\
Stanford University\\
{\tt\small mengy88@stanford.edu}
}

\maketitle
%\thispagestyle{empty}

%%%%%%%%% ABSTRACT
\begin{abstract}
   Deep artificial neural networks, trained with labeled data sets are widely used in numerous vision and robotics applications today. In terms of AI, these are called reflex models, referring to the fact that they do not self-evolve or actively adapt to environmental changes. As demand for intelligent robot control expands to many high level tasks, reinforcement learning and state based models play an increasingly important role. Herein, in computer vision and robotics domain, we study a novel approach to add reinforcement controls onto the image recognition reflex models to attain better overall performance, specifically to a wider environment range beyond what is expected of the task reflex models. Follow a common infrastructure with environment sensing and AI based modeling of self-adaptive agents, we implement multiple types of AI control agents. To the end, we provide comparative results of these agents with baseline, and an insightful analysis of their benefit to improve overall image recognition performance in real world.
\end{abstract}

%%%%%%%%% BODY TEXT
\section{Introduction}

The quality of real time images and live video feeds to image recognition tasks are often below what is expected of trained task models, in a real world scenario this can be caused by many reasons not limited to sensor limitations (e.g. lack of stabilization), movements of target or local objects, extreme weather or illumination (headlight, strong background light or low light) conditions, etc.

Most de facto deep learning models in vision perception and image recognition tasks (classification, localization, segmentation) are pretrained supervised models, it is worth studying their behavior in \textit{common} exceptional conditions; here we propose an AI adaptive solution with reinforcement learning to complement an existing pretrained reflex model to better handle a wide dynamic environment range.

%-------------------------------------------------------------------------
\subsection{Problem}

Most supervised models in image recognition are trained over a set of common images. Even though, augmentation is widely applied to create models with better generalization, they still fall short to simulate many real world scenarios. Augmentations similar to real scenes are sometimes hard or expensive to implement at training; for example images with blurriness or bad exposure are rarely used to train a deep learning model. The true challenge is that real world imagery quality is often degraded beyond that training set presents. Figure~\ref{fig:one} below shows us some well known examples from driving domain:

\begin{figure}[h]
\begin{center}
   \includegraphics[width=1.0\linewidth]{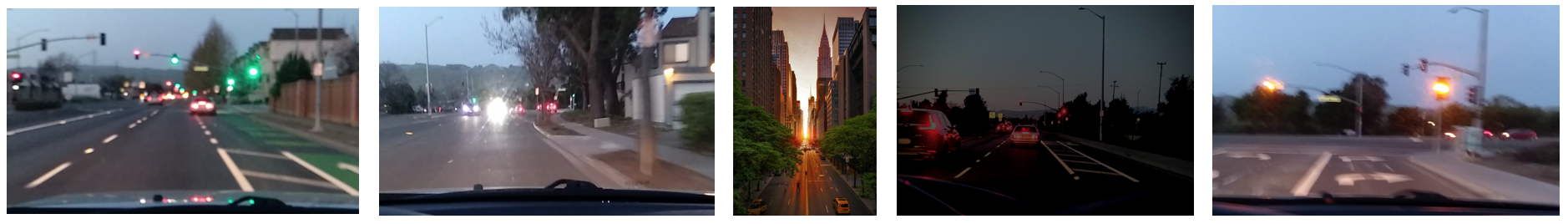}
\end{center}
   \caption{Degraded image quality by noises [blur, shaky, dark, over-exposure] to camera view at typical driving scenario.}
\label{fig:long}
\label{fig:one}
\end{figure}

Therefore, we look forward to methods that fix up imagery qualities online to complement an existing perception models.

\subsection{Motivation}

Reinforcement learning is a potential solution to many control problems, i.e. it can be used to compensate modeling loss from a fixed set of parameters or model limitations. AI in bigger scope also addresses world dynamics with adaptive state based modeling. Here we strive to add an online learner (AI based adaptive image filter agent) in front of a well defined and pretrained task (supervised object detector) network, and study the system performance end to end. The abstract idea may not be the first at large, but the work to combine and compare multiple reinforcement learning (Fast/Q-Learning ) agents with a specific DNN model for any end to end task has it's edge and novelty.

Even though, the motivation stems from Driving domain, to commence a clear study that is easier to follow, we perform the experiments over different setting with a simpler task network and smaller data sets.

%-------------------------------------------------------------------------

\section{Related work}

To the application of learning methods, many proposals exist for similar setting but different problems. To non-contextual Multi-Armed bandit (MAB) problem, Lai and Robbins ~\cite{UCBLai} and Lai ~\cite{Lai1, Lai2} showed efficient parametric solutions, proving that upper confidence bound (UCB) algorithms are optimal with minimum regret if the rewards are i.i.d. But many real MAB problems are rather contextual vs. pure random. Contextual bandits extend MAB by making the decision conditional on the state (observation) of the environment, therefore, it is widely applied in various applications. For example, Lihong Li and Wei Chu and John Langford \etal ~\cite{linucbdisjoint} used it to personalize news articles, and further in ~\cite{linucblinearpayoff} Wei Chu and Lihong Li \etal studied contextual bandits with linear payoff. On top of these, Ku-Chun Chou and Chao-Kai Chiang \etal ~\cite{prlinucb} further explored an idea to feed learning agent some pseudo rewards on non-selected arms at each action, motivated by the fact that a better choice may exist if a hypothetical reward to non-selected actions is revealed to the agent; also Xiao ~\cite{LinUCB} applied this similar idea in a Pharmacological Dose estimation problem.

In the more general front, Reinforcement Learning since Richard E. Bellman has made great progress over time, from Q-Learning ~\cite{QL} by Christopher JCH Watkins and Peter Dayan to Deep Reinforcement Learning by DeepMind and DQN ~\cite{DQN} by Hasselt, Guez and Silver, etc. We have seen consistent interests in applying RL to image recognition (related) tasks. 

In image domain, Sahba and Tizhoosh has conducted research in ~\cite{filterfusion} to image enhancement based on fusion of a number of filters using reinforcement learning scheme, their work uses a scalar reward that is determined subjectively by an end user; instead we propose a reinforcement learning agent working directly in concert with a deep learning task network, with RL control in loop, the work aims to remove an user in the loop and have the RL agent automatically pick single best filter soon at observing an input image, so that the performance of end task network is optimized. We focus on a single shot best filter in consideration of the performance over an observation period with a realistic inference cost from image ingestion to perception outcome in typical real time applications.

Sahba and Tizhoosh \etal ~\cite{reinforceSegment} further investigated a fast learning method at a time called Opposition-Based RL in image segmentation in the field of medical space. It uses an agent-based approach to optimally find the appropriate local values to help with segment tasks. The agent uses an image and its manually segmented version to take actions and gather rewards from the quality of segmented image (with manual effort in loop). It proposed Opposition-Based RL to speed up explore and exploit the solution space. But this earlier work had no exposures to recent deep learning advancement in the task space, due to an earlier time constraints.

Furuta and Inoue \etal ~\cite{fcnmsreinforcement} recently studied reinforcement learning with pixel-wise rewards (pixelRL) for image processing; there each pixel has an agent, which changes the pixel value via an action. In pixelRL, an effective learning method is proposed and it significantly improves the performance by considering not only the future states of the own pixel but also those neighbor pixels. pixelRL can be applied to some image processing tasks that require pixel-wise manipulations, but deep RL has not been applied. Author have applied the proposed method to three image processing tasks: image de-noising, image restoration, and local color enhancement with observable comparable or better performance, compared with state-of-the-art supervised learning methods.

Most recently, there are quite some new articles focus on applying attention mechanism or active localization with deep RL technique. For example, Caicedo and Lazebnik ~\cite{ActLocal} explored an active detection model for object localization using RL agent; and Chen and Wang \etal ~\cite{Atten} proposed a recurrent attention reinforcement learning to discover a sequence of attentional and informative regions for better multi-label objects recognition.

Nevertheless, the Fast/RL agents that we propose and study herein do not address direct object recognition aspects, such as optimal region proposals, etc. Instead, we look straight into rather operable legacy CV filters (soften, sharpen, whiten or darken, etc.) as possible {\bf Actions} to the control, then look forward to an automated learning process of an optimal policy, to enhance end to end image recognition task performance using any pre-allocated DNN task backbone (reflex model).

Primarily our study focuses on using Fast/RL method to propose an AI single shot agent over CV filters action space, aiming to apply most recent state of the art deep learning task models usable to much wider range of applicable environment, and for a better end to end task performance with automation. We look forward to our lightweight solution suitable to real time autonomous machinery applications.

%-------------------------------------------------------------------------

\section{Dataset, oracle and baseline}
\label{sec:dataset}

For this remaining work, we reduce to use \textbf{Flowers102} dataset, and pretrained (on Image Net) \textbf{VGG-19} based detector \textbf{D} as the task image recognition DCNN network. To construct \textbf{D}, we replace the top classifiers of \textbf{VGG-19} with custom modified classification layers fit and retrained on the Flowers dataset. At the time to implement and evaluate AI online filter agents, \textbf{D} has reached a 80.5\% classification accuracy on the test set of \textbf{Flowers102} (train/val/test split: 80\%/10\%/10\%), the network model is then frozen and not adjusted again, to provide a consistent measurement thereafter.
\\

To study an online learning agent of image filter to improve \textbf{D}'s performance on noisy images, assumption is also made so that de-noise is only done on per image basis. Current work does not address de-noise across images, nor look into inter-image noise correlations.

Three main noisy effect scenarios are chosen to investigate: darkness, whiteness and blurriness. To create a noisy image dataset for these scenarios, we create synthetic dataset by applying three different OpenCV noisy filters (darken, whiten and blur) over an original Flowers dataset. Accordingly, corresponding de-noise filters (whiten, darken and de-blur) are introduced and chosen for quality corrections that shall be used as actions by AI agent for automated image filtering. Figure~\ref{fig:two} below shows some corrected examples from these properly chosen de-noise filters, which are applied to correct blur, dark and white noisy images respectively here.

\begin{figure}[h]
  \centering
  \includegraphics[width=1.0\linewidth]{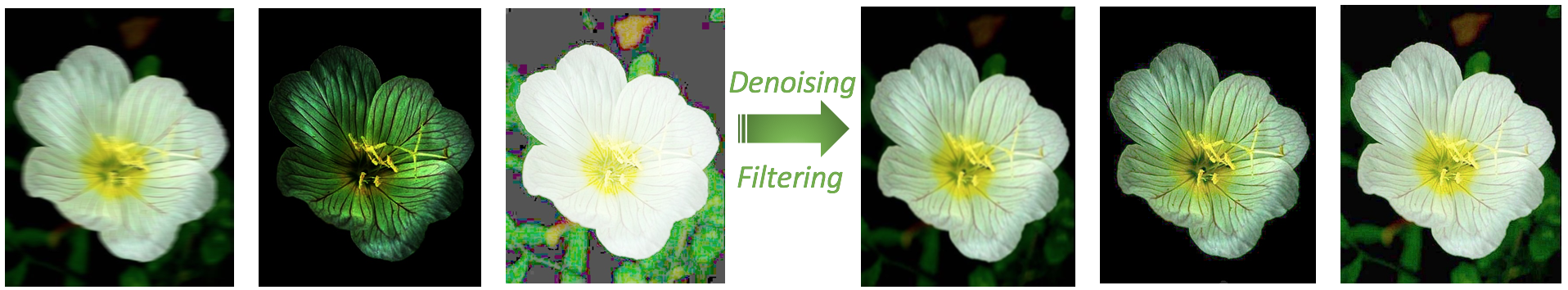}
    \caption{AI agent noise removal example results}
\label{fig:two}
\end{figure}

\subsection{Dataset, preprocess and synthetic data}
Three types of synthetic noisy image sets are generated using OpenCV noise filters below:
\begin{itemize}

\item Blur filter: blurs image by smoothing the image with a 5x5 filter kernel.
\item White filter: achieves overexposure using gamma correction of high gamma value (e.g 3.5).
\item Dark filter: simulates underexposure using a low gamma value (e.g gamma=0.2).

\end{itemize}

\subsection{de-noise filters for AI agent}
Following OpenCV de-noise filters are introduced, for online Agent use as counter actions.
\begin{itemize}
\item De-blur filter: use \verb+ cv.filter2D()+ remove blurriness with a 3x3 sharpen kernel of center value as 9 and the rest values as -1.
\item Whiten filter: increase brightness with high gamma.
\item Stronger whiten filter: increase brightness stronger with even higher gamma value.
\item Darken filter: adjust overexposure images with gamma correction.
\item Stronger darken filter: darken the image with stronger effect with even lower gamma value.
\end{itemize}

\subsection{Baseline vs. oracle performance}
Baseline and oracle are based on the result from the image classifier detector \textbf{D}, which outputs top k categorical prediction probabilities for each image. As the true label of each image is known, correctness of prediction for the target class can be computed from the softmax probability of each image at inference.

$Definition:$

$\textbf{Correct\_Softmax\_Probability}$

$= \hat{P}(label_{top1\_prediction} == label_{ground\_truth})$

\begin{itemize}
\item \textbf{Baseline}: for each image, baseline is the correct softmax probability returned by detector \textbf{D} over the noised image, without any de-noise filter being applied.
\item \textbf{Oracle}: for each image, oracle is the correct softmax probability by detector \textbf{D} over its original image in Flowers dataset, without any noise introduced or de-noise filter applied.
\end{itemize}

\par
To see that baseline and oracle are true lower and upper bounds, an offline comparison detector \textbf{D}'s result on images after applying correct de-noise filters is provided. Table 1 shows that in the average case, an AI agent produces an image correction with \textbf{D}'s prediction accuracy lies between baseline and oracle.

\begin{table}[h]
  \begin{center}
  \scalebox{0.8}{
  \begin{tabular}{ll|l}
    \hline
    \multicolumn{3}{c}{\textbf{D}'s prediction accuracy}                   \\
    \hline
    Scenario     & Without de-noise filter    & With proper de-noise filter \\
    \hline
    \hline
    Oracle images  & 67.99\%   & -     \\
    Blur images    & 59.04\%   & 63.12\%      \\
    Dark images    & 50.79\%   & 58.89\%  \\
    White images   &56.16\%    &65.27\% \\
    \hline
  \end{tabular}}
  \end{center}
  
  \caption{prediction accuracy on different image set}
  \label{baseline-oracle-table}
\end{table}

%-------------------------------------------------------------------------

\section{Methods and approaches}

An image filter agent needs to take \textbf{Actions} from \textbf{States} that it perceives from environment. Images coming from an environment is simulated by introducing random noises [in this study, images are read from stored files vs. continuous live stream]. A scalar reward function is also introduced to provide feedback to actions taken by the agent.

\subsection{States}
Agent takes action on each image and observes rewards, it is unnecessary to have its state persistent across images. In concept, this is similar to contextual multi-armed bandit. As such, agent state need to have contextual information on the current input image. Herein an agent state with four feature variables regarding image quality is defined. For each agent State $S(x_{1}, x_{2}, x_{3}, x_{4})$:
\begin{itemize}
\item \textbf{$x_{1}$}: \textbf{Variance of Laplacian} used to indicate Blurriness. This feature is quantized in scale $[0, 1, 2]$ with larger number indicates blurrier.
\item \textbf{$x_{2}$}: \textbf{Overall brightness}. This image feature is quantized in scale $[-1, 0, 1]$ by comparing with the \textbf{mean} of detector \textbf{D}'s training set (This tells the difference of real inference image's brightness with \textbf{D}'s mean expect).
\item \textbf{$x_{3}$}: mean of \textbf{V} value from \textbf{HSV} image channels. This feature describes the intensity of the image color and is quantized in scale $[0, 1, 2]$.
\item \textbf{$x_{4}$}: mean of \textbf{L} value from \textbf{HSL} image channels. This feature describes the lightness of the image color and is quantized in scale $[0, 1, 2]$.
\end{itemize}

\subsection{Action and reward}
\paragraph{Action} a total of six actions are introduced to AI agent in current study. When agent observes an input image, it applies one of the de-noise filters, or takes no action at all. Therefore, the space of \textbf{Actions} are: \textbf{[None, De-blur, Weak whiten, Strong whiten, Weak darken, Strong darken]}

\paragraph{Reward} 
for each action taken is defined upon comparison of the detector \textbf{D}'s accuracy on the filtered image with an oracle accuracy from its original image. Rewards are then quantized in a range of $[-6, 2]$.

Environment has precomputed a table that maps an oracle \verb+image_name+ to its correct softmax probability \verb+oracle_pr+ by detector \textbf{D}. Reward function then extracts correct (on target class) softmax probability \verb+denoise_pr+ of an de-noised image from AI filter agent using the same \textbf{D}. A probability drop threshold \verb+pd+ is provided so that a higher \verb+denoised_pr+ gets better rewards. For example, better reward is given if $\verb+denoise_pr+ > (\verb+oracle_pr+ - 1.0*\verb+pd+)$.

%-------------------------------------------------------------------------

\section{State diagram and architecture}
\begin{figure}[h]
  \centering
  \includegraphics[width=1.0\linewidth]{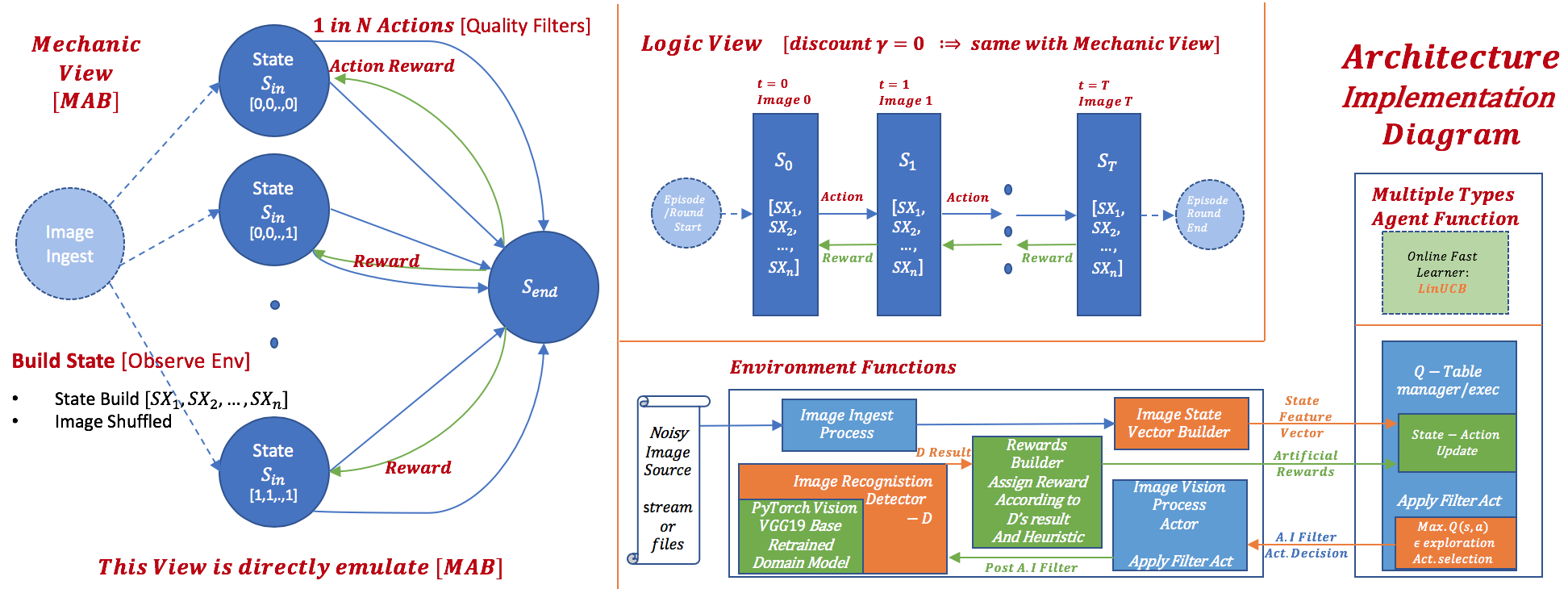}
    \caption{State diagram and implementation architecture}
\label{fig:three}
\end{figure}

With the problem as it presents, \textbf{Two} different methods are introduced to the design of an online image filter agent: MAB with \textbf{LinUCB} and RL with \textbf{Q-Learning}.

\subsection{State machine}
Figure~\ref{fig:three} left shows the state diagram mainly to emulate \textbf{MAB} in transition, while the upper central part emulates a model free \textbf{Markov} chain in transition, that is later solved with \textbf{Q-Learning}.
 
\begin{itemize}
\item \textbf{MAB} (Multi-Armed Bandit): as a logical choice, reward is immediately assigned at picking an image filter each time. To have a learnable agent, fast online learning algorithm, such as Contextual Linear Bandit is implemented. The left half of Figure~\ref{fig:three} shows a mechanic view of this approach.
\item \textbf{Q-Learning}: When the task of online image filtering is considered as a sequence of state \textbf{s}, action \textbf {a}, reward \textbf {r}, next state \textbf{$s'$}, with managable action and state space, it may be solved with \textbf{Q-Learning} agent for an optimal policy from experience. The middle part of Figure~\ref{fig:three} does show the logic view of state transitions, where each state presents an observed image with state variables, and action presents an de-noise filter in application. \textbf{Note} when discounted factor $\gamma=0$, this sequential state diagram equally reverts back to the \textbf{MAB} state view to the left.
\end{itemize}

\subsection{Implementation}
For the benefit of modularity and reusability, we strive to design a common framework to support:
\begin{itemize}
\item multiple types of \textbf{AI agent classes} with different online learning algorithms.
\item multiple \textbf{environment functions} for an easy extension to different task network or deep reflex models \textbf{D}, different state sensing mechanism and reward emitting scheme.
\end{itemize}

In terms of design, followings are implemented within current study, as demonstrated in Figure~\ref{fig:three}:
\begin{itemize}
\item \textbf{Environment Functions}
    \begin{itemize}
        \item \textbf{State sensor}: this function ingests incoming noisy image, and build up state upon its imagery quality. It outputs state vector (RL) or feature variables (LinUCB). In case of $(s, a, r, s')$ sequence, it may extract a next state $s'$ by prefetching next image.
        \item \textbf{Action actor}: this function realizes the \textbf{Action} determined by AI agent. It applies designated image filter, and passes post filtered image onto the next task network - \textbf{D}.
        \item \textbf{Tasks runner}: this is the image recognition task network - \textbf{D}, it uses a reflex deep learning model, and it can be swapped with another task networks on demand.
        \item \textbf{Reward emitter}: this is the reward extraction function, it emits an action \textbf{reward} from \textbf{D}'s task performance observation on a de-noised image. This function is key to an overall performance of a reinforcement learning control loop. It may be implemented in many different ways, and its choice is also an active field in research and practice (e.g. how to come up with a reward with minimum prior or oracle to start with). In this study, situation is simplified to assume an available oracle (original) images, therefore reward is given according to an observed difference via \textbf{D}'s performance.
    \end{itemize}

\item \textbf{Agent Functions}
    \begin{itemize}
        \item \textbf{LinUCB}: fast online learning agent with Contextual Linear Bandit algorithm, it uses linear pay-off mechanism, takes state variable $s$ as feature vector $X$
        \item \textbf{Q-Learning}
            \begin{itemize}
                \item \textbf{Q-Learning with $\gamma=0$}: ignore the next image/state $s'$
                \item \textbf{Q-Learning with $\gamma>0$}: consider discounted future rewards, via prefetch and lookahead of the next image/state $s'$
            \end{itemize}
    \end{itemize}
\end{itemize}

%-------------------------------------------------------------------------

\section{Models and Experimental Results}
\label{sec:result}

To design an AI agent to learn optimal image filtering from past experience, the problem can be framed as a reinforcement online learning problem with specific environments introduced above. This is approached with \textbf{two} different \textbf{RL} models of algorithmic types in current study.

\subsection{LinUCB}
In the MAB problem setting, quality features of an incoming image is currently observed as $X(x_{1}, x_{2}, x_{3}, x_{4})$, which is used to define the context. Therefore, agent can use Contextual Linear Bandit to make prediction conditional on the state (observation of $X$) of an observed environment.

Combine this with a parameterized linear payoff (rewards): $r_{t, a}=X_{t,a}^{T}\theta_{a}^{*}$ (parameter $\theta_{a}$ is learnable), linear disjoint (that each arm/filter has its own parameter) LinUCB can be used for the solution.

\subsubsection{Algorithm}
% Algorithm bullet point

According to ~\cite{ucblecture}, ~\cite{linucblinearpayoff} and ~\cite{linucbdisjoint} we can have the disjoint arm algorithm of \textbf{LinUCB} written out below:\

\begin{figure}[h]
  \centering
    \includegraphics[width=1.0\linewidth]{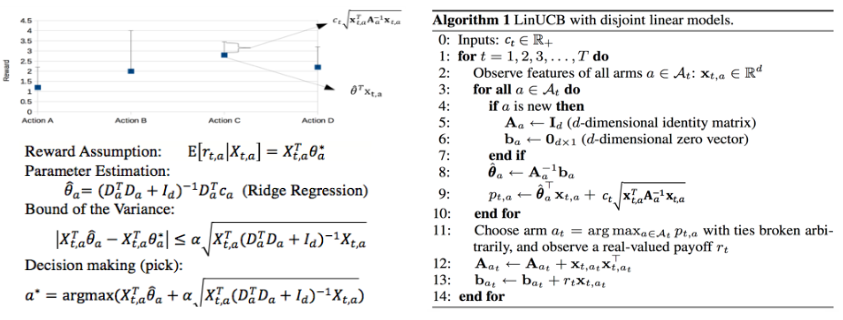}
    \caption{LinUCB algorithm}
    \label{fig:four}
\end{figure}

In Figure~\ref{fig:four} pseudo code , $c_{t}$=$\alpha$ is parameter to balance exploitation vs. exploration; $X_{t,a}$=$X_{t}$ is feature vector seen by agent at time $t$ (not to consider per \textbf{Action} feature set); $r_{t}$ is reward agent receives at time $t$.

\subsubsection{LinUCB results}

Here MAB is implemented with \textbf{LinUCB class}. At execution, it interacts with a common \textbf{Environment} class instance (with \textbf{D} for task inference) for an end to end online learning: it uses previous observation $X_{t}$, action $a_{t}$ and reward $r_{t}$ to pick an optimal image filter, for current iteration to apply on incoming noisy image.

\begin{figure}[h]
  \centering
    \includegraphics[width=1.0\linewidth]{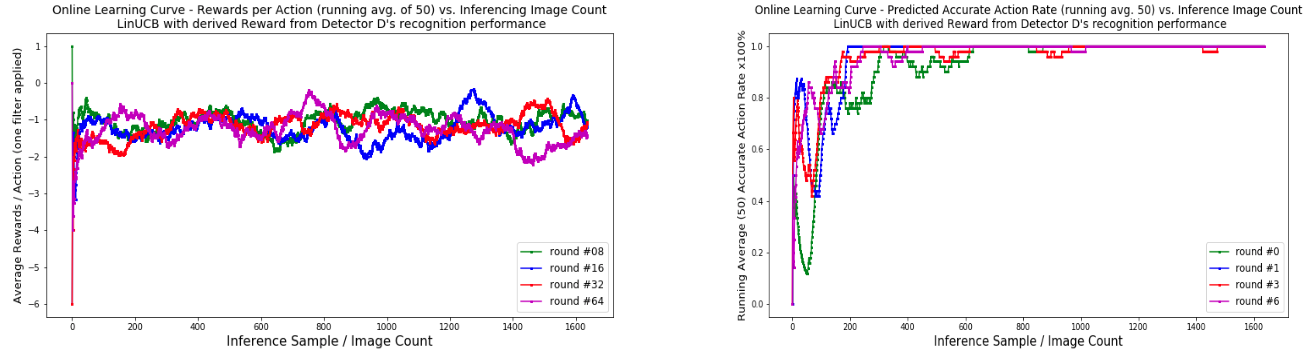}
    \caption{Left: LinUCB learning curve - running average reward per action vs. inference image count. \newline \hspace{13mm}Right: LinUCB learning curve - running average accurate action rate vs. inference image count.}
    \label{fig:five}
\end{figure}

Figure~\ref{fig:five} shows the LinUCB Agent performance and learning curves.
\newline
\newline
\textbf{LinUCB} Agent \textbf{results} and \textbf{analysis}: 
 
\begin{itemize}
\item In terms of average \textbf{reward per action}, LinUCB Agent learns quickly to reach a reasonable level (s.t. not much performance degradation compare to oracle according to \textbf{D}'s predictive probability) in just tens of initial iterations for each round. (as annotated in left plots above)
\item In terms of average \textbf{prediction accuracy} (accurate filter taken), LinUCB Agent gradually reaches 80.0\%+ accuracy in ~100s iterations, and then gets to near ~100.0\% accuracy in a few hundreds of iterations. (observations from random rounds shows a similar result; images are shuffled in ingestion pipeline at each round)
\end{itemize}

\subsection{Q-Learning}
In this problem setting, there is not much need to care about the state transitions (i.e. how current image may lead to next one to come). Therefore, as model free method, Q-Learning suits the needs well.

Due to the short episode (considerably as MAB when $\gamma=0$) $len=1$, relative small \textbf{state space}: $size=|3|^4$ and small \textbf{action space}: $size=6$, the solution is approachable with a simpler \textbf{Q-Table}, versus a function approximator Q-Network or complex \textbf{DQN} as Hasselt and David Silver \etal had proposed in ~\cite{DQN} unless the state-action space explodes notably large.

At each iteration $t$, the \textbf{QLearning Agent} takes on a state from an observation $S_{t}$, picks an action $a_{t}$  (filter) via $\epsilon$-$greedy$ algorithm, upon receiving reward $r_{t}$ from \textbf{D} via Environment, it updates the tabular $\hat{Q}_{opt}(S_{t}, a_{t})$ with \textbf{learning rate $\eta$} with or without the maximum $\hat{Q}_{opt}$ from the next state $s'$ depending on the discounted factor $\gamma$.

\subsubsection{Algorithm: update on each $(s, a, r, s')$}
% This is Q-learning equations:
{
\small
\begin{equation}
\resizebox{.91\hsize}{!}{
$\hat{Q}_{opt}(s, a) \leftarrow (1 - \eta )\hat{Q}_{opt}(s, a) + \eta[r + \gamma\max_{a' \in Actions(s')}
\hat{Q}_{opt}(s', a')]$}
\end{equation}
}
In this application, as a special case, when there is no need to \textbf{track cross-frames} image quality correlation for the learning agent, it is okay to assume $V_{S_{end}} = 0$ (Mechanic view Figure~\ref{fig:three}) or equivalently $\gamma=0$ (Logic view Figure~\ref{fig:three}), so a simplified $\hat{Q}_{opt} $ update equation is just below:

%simplified Q-learning equation:
{
\small
\begin{equation}
   \hat{Q}_{opt}(s, a) \leftarrow \hat{Q}_{opt}(s, a) + \eta[r - 
\hat{Q}_{opt}(s, a)]
\end{equation}
}

\subsubsection{Q-Learning result analysis}

Using prefetch or lookahead (next image), it is possible to profile Q-Learning Agent overall behaviour with different discounted factor $\gamma$. At doing this, \textbf{Q-Table} is not reset between rounds (each round is a full iteration of all learning and test noisy images) to have Q-Matrix filling up overtime.

\begin{figure}[h]
  \centering
    \includegraphics[width=1.0\linewidth]{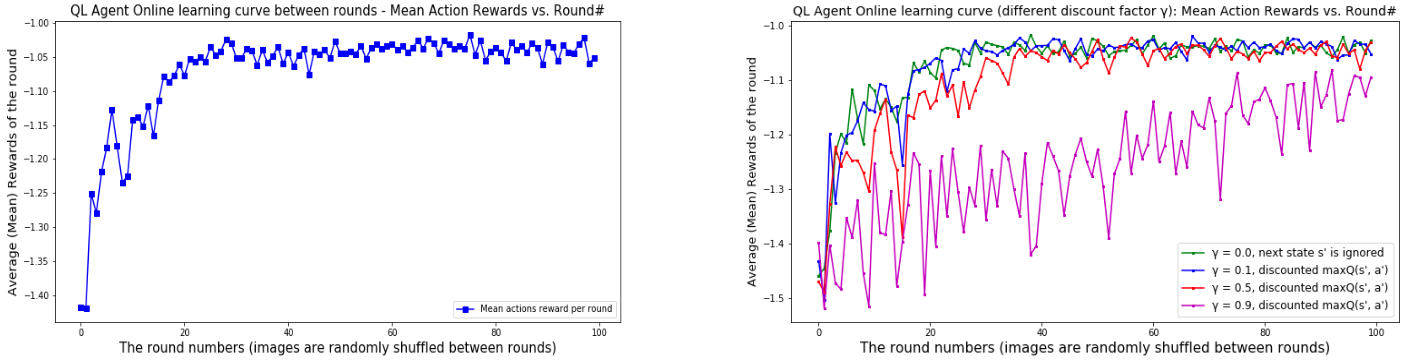}
    \caption{Left: Q-learning curve of mean action rewards per round from all rounds.  \newline \hspace{27mm}Right: Q-learning curve of mean action rewards per round from all rounds with different $\gamma$}
    \label{fig:six}
\end{figure}

Demonstration from Figure~\ref{fig:six}.
\begin{itemize}
\item \textbf{Foremost}, the Q-Learning agent is truly \textbf{effective}, left diagram above is generated with \textbf{vanilla QLAgent} using simplified \textbf{one-hop (state)} update \textbf{equation (2)}, it is clear that the average action reward per round rises upwards as the round continues. (reward resides in negative range, due to a less generous rewarding scheme by design.)
\item Diagram to the right shows the same curves from different discount $\gamma$ values. They are generated from \textbf{full QLAgent} with next state $s'$ prefetch and update \textbf{equation (1)}. As a sanity check, curve in left diagram reconciles with the $\gamma=0$ curve in diagram to the right.
\end{itemize}

It is observed that larger $\gamma$ does not help in current problem setting, \textbf{reason} being our synthetic noisy images are randomly shuffled vs. being continuous and potentially correlated from a video stream, therefore incorporating non-naturally correlated next state $s'$ may not help with Q-Learning's convergence at all. In remaining study \textbf{vanilla QLAgent} is used except explicitly mentioned otherwise.    \textbf{Q-Learning} Agent \textbf{performance}: 
\begin{figure}[h]
  \centering
    \includegraphics[width=1.0\linewidth]{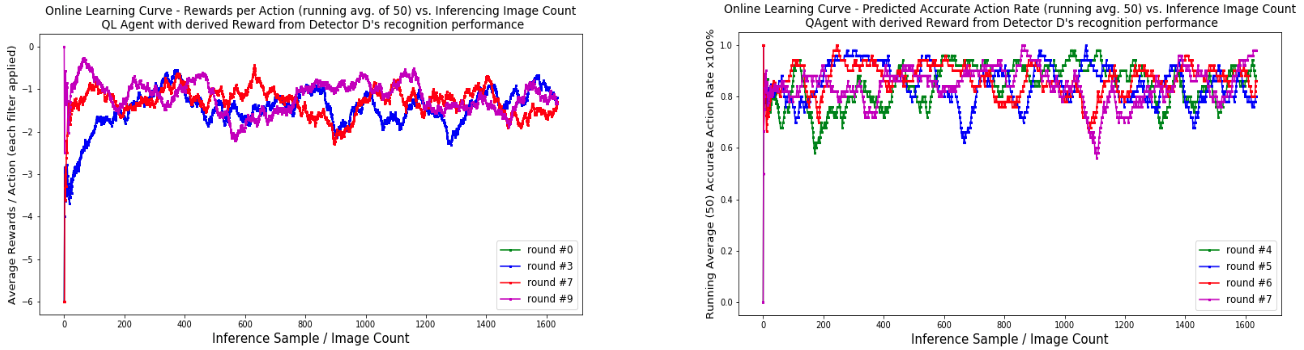}
    \caption{Left: Q-learning curve - running average reward per action vs. inference image count.  \newline \hspace{13mm}Right: Q-learning curve - running average accurate action rate vs. inference image count.}
    \label{fig:seven}
\end{figure}

Figure~\ref{fig:seven} shows the Q-learning Agent performance and learning curves.

\begin{itemize}
\item Regarding average \textbf{reward per action}, (vanilla) Q-Learning Agent learns relatively quick to reach preferable level, in around hundred of iterations of each round. (seen above left)
\item In terms of average \textbf{prediction accuracy} (correct filter taken), Q-Learning Agent reaches 80.0\%+ accuracy range relatively fast in just tens of iterations, but it is not so stable at near to ~100.0\% accuracy for many iterations within a round. (due to non-optimal parameter tuning of $\epsilon$-$greedy$ for exploration vs. exploitation, limited iterations [image counts] per round, etc.)
\end{itemize}

\textbf{Q-Learning} Agent \textbf{online performance} and analytic comparison with \textbf{oracle, baseline}: 
\begin{figure}[h]
  \centering
    \includegraphics[width=1.0\linewidth]{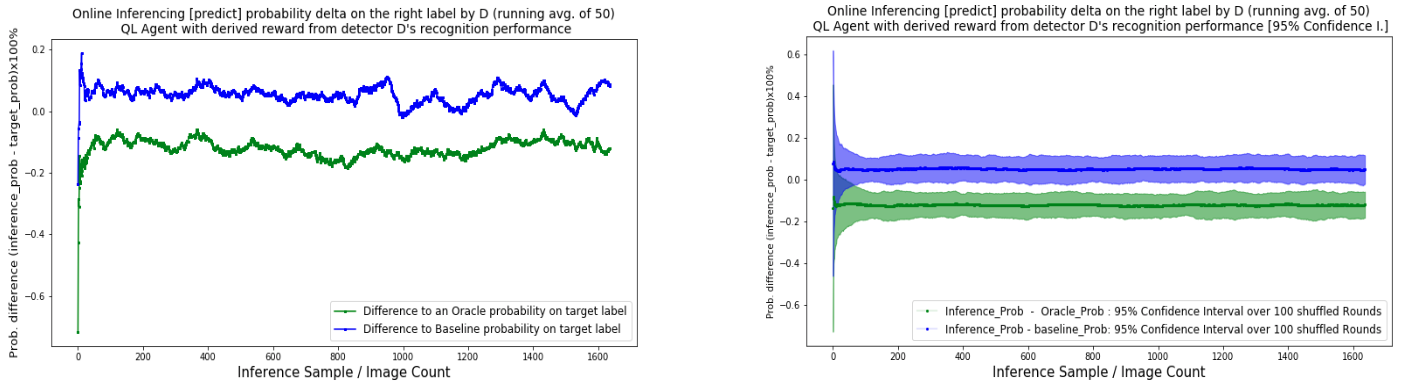}
    \caption{Left: inference probability (on correct label) difference from QLAgent's policy with baseline, oracle respectively.  \newline Right: inference probability (on label) difference from QLAgent's policy with baseline, oracle respectively with 95\% Confidence Interval}
    \label{fig:eight}
\end{figure}

Figure~\ref{fig:eight} shows the difference of \textbf{D}'s softmax probability on the correct label between de-noised images by image correction QLAgent with pre-measured predictive probability over oracle (original) or baseline (noisy) images. It demonstrates several key and interesting informative points:
\begin{itemize}
\item A consistent range of 15 $~$ 20\% difference in softmax probability is seen between \textbf{oracle} (green) and \textbf{baseline set} (blue), when the \textbf{D}'s inference result on QLAgent corrected (de-noise) image is used as a middle reference. This reconciles with the Table 1 data in 3.3
\item In average, about 10\% softmax probability drop comparing to \textbf{oracle} image set; and about 5\% softmax probability increase (on the correct label) comparing to \textbf{baseline} image set. It is a strong evidence that the online \textbf{AI Agent} plays an \textbf{effective} role in correcting noisy images to attain better image recognition overall performance in the end, and it learns an optimal policy to act purely \textbf{online}.
\end{itemize}

\subsection{Conclusion and analysis}
\textbf{Conclusion}: \textbf{TWO} different AI image filtering agents are proposed in the problem setting, they are both \textbf{effective} in enhancing image quality online to improve recognition performance according to our experiments. To gain further insight, side by side comparison between the two agents is also provided.

\textbf{Q-Learning vs. LinUCB} Agents: prediction accuracy of right \textbf{(filter)} with 95\% confidence interval 

\begin{figure}[h]
  \centering
  \includegraphics[width=1.0\linewidth]{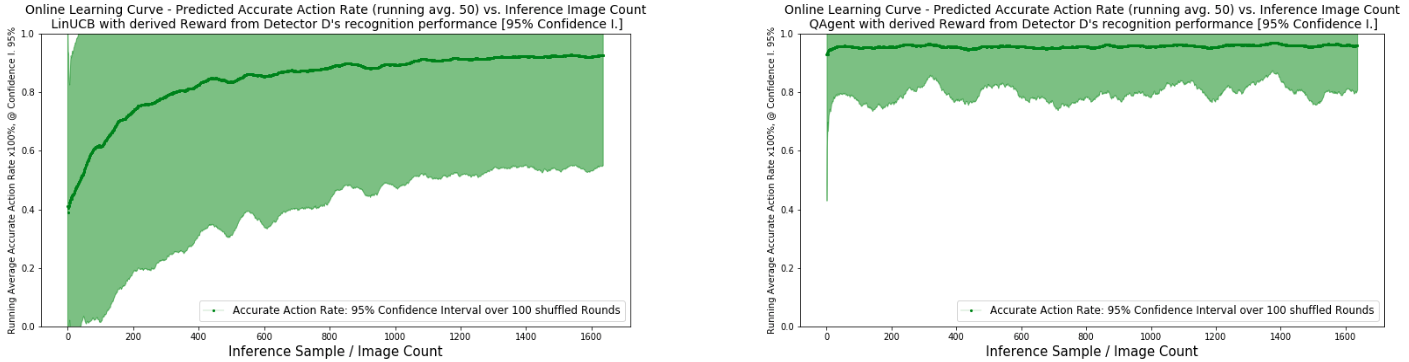}
    \caption{Left: LinUCB running average of predicted accurate action rate vs. inference count with 95\% confidence interval.  \newline \hspace{8mm}Right: QLAgent running average of predicted accurate action vs. inference count with 95\% confidence interval.}
    \label{fig:nine}
\end{figure}

Figure~\ref{fig:nine} shows a different confidence band between our LinUCB and QLAgent implementation.

\begin{itemize}
\item To both agents, average prediction accuracy (rate of right filter being selected) rises above 90\% as inference count increases. QLAgent quickly attains even higher accuracy of 95\%+.
\item \textbf{LinUCB} Agent learns and improves slower than \textbf{Q-Learning} Agent in average, with larger variation (confidence band) in observation. This difference is also seen in Figure below:
\end{itemize}

\begin{figure}[h]
  \centering
  \includegraphics[width=1.0\linewidth]{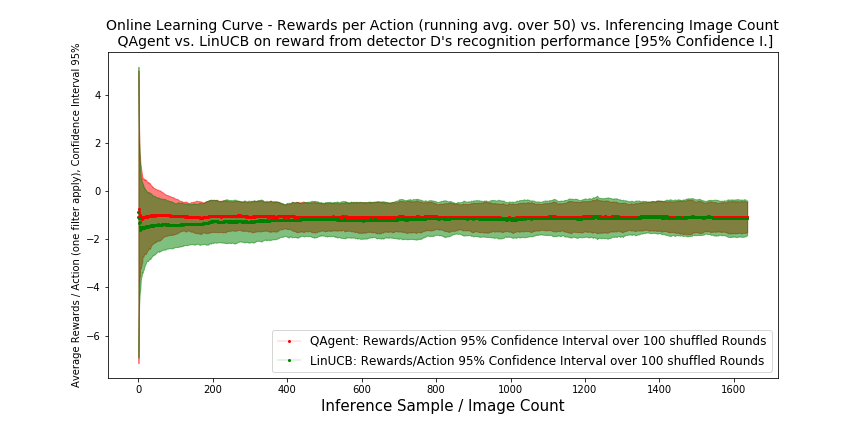}
    \caption{QLAgent vs. LinUCB - running average rewards per action vs. inference count with 95\% confidence interval.}
    \label{fig:ten}
\end{figure}

Figure~\ref{fig:ten}. shows that \textbf{QLAgent} (red) moderately outperforms \textbf{LinUCB} agent (green) in average reward per action, in the experiments of current setting. \textbf{Analysis}: One reason of the difference is lack of LinUCB tuning, since only a default parameter $\alpha={c_t}=1$ (in balancing exploitation vs. exploration) is used in current study. Another possibility is that a small Q-Learning learning rate $\eta=0.002$ helped to smooth out some level of instantaneous variation from $\hat{Q}_{opt}$ update.

%-------------------------------------------------------------------------

\section{Conclusion}
\label{sec:conclusion}

Automated effective image filtering for performance is a challenge and interesting task, rule based approach may be effective in specific situation of certain domain, but it falls short to perform well in a generic scenario. We believe an online learning based fast reinforcement learning agent will exceed traditional methods in terms of flexibility and generalization, especially when it is fused with a task oriented deep reflex DNN model. In this study, we explored premier CV (ISP) image filters (action), AI, Reinforcement Learning (control) and Deep Learning (reward) for a complete end to end learning system. Our result looks promising. On the AI online image filter function, we have implemented both Q-Learning and LinUCB agents working with a complex but common environment function, to learn proper image correction filter towards an overall better end to end image recognition by a common deep learning detection module. In our experiments of this problem setting, LinUCB agent learns quicker, but Q-agent shows a better result in terms of accuracy and convergence. To conclude, we have demonstrated with both approaches the benefits to fuse an online RL fast learning agent with a Deep Learning reflex network to gain an overall task performance.

\paragraph{Future discussion} here are some future directions, extended from current study:
\begin{itemize}
\item Extend \textbf{State} space: explore more effective CV features with regarding to imagery qualities, with our flexible framework, this is straight forward to do.
\item Extend \textbf{Action} space: explore more effective image quality correction filters as new actions; it's quite extensible as well, with our flexible framework.
\item Realistic \textbf{reward scheme} (heuristic) for real world inference: in real situations, extended mechanism in real time rewarding is needed; this can be done in several ways, including but not limited to: side-channel or outbound reference, data fusion with different sources.
\item Incorporate cross-frames image quality tracking for corresponding corrections, look into the performance of our agents on highly correlated images from continuous video feeds.
\end{itemize}

%-------------------------------------------------------------------------

{\small
\bibliographystyle{ieee_fullname}
\bibliography{egbib}
}

\end{document}